# Policy Gradients with Variance Related Risk Criteria

**Aviv Tamar**                                                          AVIVT@TX.TECHNION.AC.IL
**Dotan Di Castro**                                                      DOT@TX.TECHNION.AC.IL
**Shie Mannor**                                                         SHIE@EE.TECHNION.AC.IL
Department of Electrical Engineering, The Technion - Israel Institute of Technology, Haifa, Israel 32000

## Abstract

Managing risk in dynamic decision problems is of cardinal importance in many fields such as finance and process control. The most common approach to defining risk is through various variance related criteria such as the Sharpe Ratio or the standard deviation adjusted reward. It is known that optimizing many of the variance related risk criteria is NP-hard. In this paper we devise a framework for local policy gradient style algorithms for reinforcement learning for variance related criteria. Our starting point is a new formula for the variance of the cost-to-go in episodic tasks. Using this formula we develop policy gradient algorithms for criteria that involve both the expected cost and the variance of the cost. We prove the convergence of these algorithms to local minima and demonstrate their applicability in a portfolio planning problem.

## 1. Introduction

In both Reinforcement Learning (RL; Bertsekas & Tsitsiklis, 1996) and planning in Markov Decision Processes (MDPs; Puterman, 1994), the typical objective is to maximize the cumulative (possibly discounted) expected reward, denoted by $J$. When the model's parameters are known, several well-established and efficient optimization algorithms are known. When the model parameters are not known, learning is needed and there are several algorithmic frameworks that solve the learning problem efficiently, at least when the model is finite. In many applications, however, the decision maker is also interested in minimizing some form of *risk* of the policy. By risk, we mean reward criteria that take into account not only the expected reward, but also some additional statistics of the total reward such as its variance, its Value at Risk, etc. (Luenberger, 1998). Risk can be measured with respect to two types of uncertainties. The first type, termed *parametric uncertainty* is related to the imperfect knowledge of the problem parameters. The second type, termed *inherent uncertainty* is related to the stochastic nature of the system (random reward and transition function). Both types of uncertainties can be important, depending on the application at hand.

Risk aware decision making is important both in planning and in learning. Two prominent examples are in finance and process control. In financial decision making, a popular performance criterion is the *Sharpe Ratio* (SR; Sharpe, 1966) – the ratio between the expected profit and its standard deviation. This measure is so popular that it is one of the reported metrics each mutual fund reports annually. When deciding how to allocate a portfolio both types of uncertainties are important: the decision maker does not know the model parameters or the actual realization of the market behavior. In process control as well, both uncertainties are essential and a robust optimization framework (Nilim & El Ghaoui, 2005) is often adopted to overcome imperfect knowledge of the parameters and uncertainty in the transitions. In this paper we focus on inherent uncertainty and use learning to mitigate parametric uncertainty.

The topic of risk-aware decision making has been of interest for quite a long time, and several frameworks for incorporating risk into decision making have been suggested. In the context of MDPs and addressing inherent uncertainty, Howard & Matheson (1972) proposed to use an exponential utility function, where the factor of the exponent controls the risk sensitivity. Another approach considers the *percentile* performance criterion (Filar et al., 1995), in which the average reward has to exceed some value with a given probability. Addressing parameter uncertainty has been done within

---

the Bayesian framework (where a prior is assumed on the unknown parameters, see Poupart et al., 2006) or within the robust MDP framework (where a worst-case approach is taken over the parameters inside an uncertainty set). Much less work has been done on risk sensitive criteria within the RL framework, with a notable exception of Borkar & Meyn (2002) who considered exponential utility functions and of Geibel & Wysotzki (2005) who considered models where some states are "error states," representing a bad or even catastrophic outcome.

In this work we consider an RL setup and focus on risk measures that involve the *variance of the cumulative reward*, denoted by $V$. Typical performance criteria that fall under this definition include

(a) Maximize $J$ s.t. $V \leq c$

(b) Minimize $V$ s.t. $J \geq c$

(c) Maximize the Sharpe Ratio: $J/\sqrt{V}$

(d) Maximize $J - c\sqrt{V}$

The rationale behind our choice of risk measure is that these performance criteria, such as the SR mentioned above, are being used in practice. Moreover, it seems that human decision makers understand how to use variance well, and that exponential utility functions require determining the exponent coefficient which is non-intuitive.

Variance-based risk criteria, however, are computationally demanding. It has long been recognized (Sobel, 1982) that optimization problems such as (a) are not amenable to standard dynamic programming techniques. Furthermore, Mannor & Tsitsiklis have shown that even when the MDP's parameters are known, many of these problems are computationally intractable, and some are not even approximable. This is not surprising given that other risk related criteria such as percentile optimization are also known to be hard except in special cases.

Despite these somewhat discouraging results, in this work we show that this important problem may be tackled successfully, by considering policy gradient type algorithms that optimize the problem *locally*. We present a framework for dealing with performance criteria that include the variance of the cumulative reward. Our approach is based on a new fundamental result for the variance of episodic tasks. Previous work by Sobel, 1982 presented similar equations for the infinite horizon discounted case, however, the importance of our result is that the episodic setup allows us to derive policy gradient type *algorithms*. We present both model-based and model-free algorithms for solving problems (a) and (c), and prove that they converge. Extension of our algorithms to other performance criteria such as (b) and (d) listed above is immediate. The effectiveness of our approach is further demonstrated numerically in a risk sensitive portfolio management problem.

## 2. Framework and Background

In this section we present the framework considered in this work and explain the difficulty in mean-variance optimization.

### 2.1. Definitions and Framework

We consider an agent interacting with an unknown environment that is modeled by an MDP in discrete time with a finite state set $X \triangleq \{1, \ldots, n\}$ and finite action set $U \triangleq \{1, \ldots, m\}$. Each selected action $u \in U$ at a state $x \in X$ determines a stochastic transition to the next state $y \in X$ with a probability $P_u(y|x)$.

For each state $x$ the agent receives a corresponding reward $r(x)$ that is bounded and depends only on the current state[1]. The agent maintains a parameterized *policy function* that is in general a probabilistic function, denoted by $\mu_\theta(u|x)$, mapping a state $x \in X$ into a probability distribution over the controls $U$. The parameter $\theta \in \mathbb{R}^{K_\theta}$ is a tunable parameter, and we assume that $\mu_\theta(u|x)$ is a differentiable function w.r.t. $\theta$. Note that for different values of $\theta$, different probability distributions over $U$ are associated for each $x \in X$. We denote by $x_0, u_0, r_0, x_1, u_1, r_1, \ldots$ a state-action-reward trajectory where the subindex specifies time. For notational easiness, we define $x_i^k$, $u_i^k$, and $r_i^k$ to be $x_i \ldots, x_k,\ u_i \ldots, u_k,$ and $r_i \ldots, r_k$, respectively, and $R_i^k$ to be the cumulative reward along the trajectory $R_i^k = \sum_{j=i}^{k} r_j$.

Under each policy induced by $\mu_\theta(u|x)$, the environment and the agent induce together a Markovian transition function, denoted by $P_\theta(y|x)$, satisfying $P_\theta(y|x) = \sum_u \mu_\theta(u|x) P_u(y|x)$. The following assumption will be valid throughout the rest of the paper.

**Assumption 2.1.** Under all policies, the induced Markov chain $P_\theta$ is ergodic, i.e., aperiodic, recurrent, and irreducible.

Under assumption 2.1 the Markovian transition function $P_\theta(y|x)$ induces a stationary distribution over the state space $X$, denoted by $\pi_\theta$. We denote by $\mathbb{E}_\theta[\cdot]$ and $\mathrm{Var}_\theta[\cdot]$ to be the expectation and variance operators w.r.t. the measure $P_\theta(y|x)$.

---

[1]Generalizing the results presented here to the case where the reward depends on the state and the action rewards is straightforward.

There are several performance criteria investigated in the RL literature that differ mainly on their time horizon and the treatment of future rewards (Bertsekas & Tsitsiklis, 1996). One popular criterion is the *average reward* defined by $\eta_\theta = \sum_x \pi_\theta(x) r(x)$. Under this criterion, the agent's goal is to find the parameter $\theta$ that maximizes $\eta_\theta$. One appealing property of this criterion is the possibility of obtaining estimates of $\nabla \eta_\theta$ from simulated trajectories efficiently, which leads to a class of stochastic gradient type algorithms known as *policy gradient* algorithms. In this work, we also follow the policy gradient approach, but focus on the mean-variance tradeoff. While one can consider the tradeoff between $\eta_\theta$ and $\mathrm{Var}_\pi[r(x)]$, defined as the variance w.r.t the measure $\pi_\theta$, these expressions are not sensitive to the trajectory but only to the induced stationary distribution, and represent the per-round variability.

Consequently, we focus on the finite horizon case, also known as the episodic case, that is important in many applications. Assume (without lost of generality) that $x^*$ is some recurrent state for all policies and let $\tau \triangleq \min\{k > 0 | x_k = x^*\}$ denote the first passage time to $x^*$.

Let the random variable $B$ denote the accumulated reward along the trajectory terminating at the recurrent state $x^*$

$$B \triangleq \sum_{k=0}^{\tau-1} r(x_k). \tag{1}$$

Clearly, it is desirable to choose a policy for which $B$ is large in some sense [2]. In this work, we are interested in the mean-variance tradeoff in $B$.

We define the *value function* as

$$J(x) \triangleq \mathbb{E}_\theta[B|x_0 = x], \quad, x = 1, \ldots, n, \tag{2}$$

and the *trajectory variance function* as

$$V(x) \triangleq \mathrm{Var}_\pi[B|x_0 = x], \quad, x = 1, \ldots, n.$$

Note that the dependence of $J(x)$ and $V(x)$ on $\theta$ is suppressed in notation.

The questions explored in this work are the following stochastic optimization problems:

(a) The constrained trajectory-variance problem:

$$\max_\theta J(x^*) \quad \text{s.t.} \quad V(x^*) \leq b, \tag{3}$$

---

[2]Note that finite horizon MDPs can be formulated as a special case of (1).

where $b$ is some positive value.

(b) The maximal SR problem:

$$\max_\theta S(x^*) \triangleq \frac{J(x^*)}{\sqrt{V(x^*)}}. \tag{4}$$

In order for these problems to be well defined, we make the following assumption:

**Assumption 2.2.** Under all policies $J(x^*)$ and $V(x^*)$ are bounded.

For the SR problem we also require the following:

**Assumption 2.3.** We have $V(x^*) > \epsilon$ for some $\epsilon > 0$.

In the next subsection we discuss the challenges involved in solving problems (3) and (4), which motivate our gradient based approach.

## 2.2. The Challenges of Trajectory-Variance Problems

As was already recognized by Sobel (1982), optimizing the mean-variance tradeoff in MDPs cannot be solved using traditional dynamic programming methods such as policy iteration. Mannor & Tsitsiklis showed that for the case of a finite horizon $T$, in general, solving problem (3) is hard and is equivalent to solving the subset-sum problem. Since our case can be seen as a generalization of a finite horizon problem, (3) is a hard problem as well. One reason for the hardness of the problem is that, as suggested by Mannor & Tsitsiklis, the underlying optimization problem is not necessarily convex. In the following, we give an example where the set of all $(J(x^*), V(x^*))$ pairs spanned by all possible policies is not convex.

Consider the following symmetric deterministic MDP with 8 states $X = \{x^*, x_{1a}, x_{1b}, x_{2a}, x_{2b}, x_{2c}, x_{2d}, t\}$, and two actions $U = \{u_1, u_2\}$. The reward is equal to 1 or $-1$ when action $u_1$ or $u_2$ are chosen, respectively. The MDP is sketched in Figure 1, left pane. We consider a set of random policies parameterized by $\theta_1 \in [0, 1]$ and $\theta_2 \in [0, 1]$, such that $\mu(u_1|x^*) = \theta_1$ and $\mu(u_1|x_{1a}) = \mu(u_1|x_{1b}) = \theta_2$.

Now, we can achieve $J(x^*) \in \{-2, 0, 2\}$ with zero variance if we choose $(\theta_1, \theta_2) \in \{(0,0), (0,1), (1,0), (1,1)\}$, i.e., only with the deterministic policies. Any $-2 < J(x^*) < 2$, $J(x^*) \neq 0$, can be achieved but only with a random policy, i.e., with some variance. Thus, the region is not convex. The achievable $(J(x^*), V(x^*))$ pairs are depicted in the right pane of Figure 1.

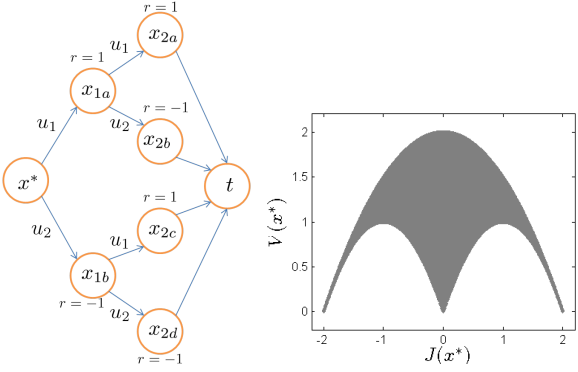

*Figure 1. (left)* A diagram of the MDP considered in Section 2.2. *(right)* A phase plane describing the non-convex nature of $J(x^*) - V(x^*)$ optimization.

## 3. Formulae for the Trajectory Variance and its Gradient

In this section we present formulae for the mean and variance of the cumulated reward between visits to the recurrent state. The key point in our approach is the following observation. By definition (1), a transition to $x^*$ always terminates the accumulation in $B$ and does not change its value. Therefore, the following Bellman like equation can be written for the value function

$$J(x) = r(x) + \sum_{y \neq x^*} P_\theta(y|x)J(y) \quad x = 1, \ldots, n. \quad (5)$$

Similar equations can be written for the trajectory variance, and in the following lemma we show that these equations are solvable, yielding expressions for $J$ and $V$.

**Proposition 3.1.** *Let $P$ be a stochastic matrix corresponding to a policy satisfying Assumption 2.1, where its $(i,j)$-th entry is the transition from state $i$ to state $j$. Define $P'$ to be a matrix equal to $P$ except that the column corresponding to state $x^*$ is zeroed (i.e., $P'(i, x^*) = 0$ for $i = 1, \ldots, n$). Then,*

*(a) the matrix $I - P'$ is invertible;*

*(b) $J = (I - P')^{-1}r$;*

*(c) $V = (I - P')^{-1}\rho,$*

*where $\rho \in \mathbb{R}^n$ and*

$$\rho(x) = \sum_y P'(y|x)J(y)^2 - \left(\sum_y P'(y|x)J(y)\right)^2.$$

*Proof.* (a) Consider an equivalent Stochastic Shortest Path (SSP) problem where $x^*$ is the termination state.

The corresponding transition matrix $P_{\text{ssp}}$ is defined by $P_{\text{ssp}}(i,j) = P(i,j)$ for $i \neq x^*$, $P_{\text{ssp}}(x^*,j) = 0$ for $j \neq x^*$, and $P_{\text{ssp}}(x^*,x^*) = 1$. Furthermore, let $P^* \in \mathbb{R}^{n-1 \times n-1}$ denote the matrix $P$ with the $x^*$'th row and column removed, which is also the transition matrix of the SSP problem without the terminal state. By the irreducibility of $P$ in Assumption 2.1 $P_{\text{ssp}}$ is *proper*, and by proposition 2.2.1 in (Bertsekas, 2006) we have that $I - P^*$ is invertible.

Finally, observe that by the definition of $P'$ we have

$$\det(I - P') = \det(I - P^*),$$

thus, $\det(I - P') \neq 0$.

(b) Choose $x \in \{1, \ldots, n\}$. Then,

$$J(x) = r(x) + \sum_{y \neq x^*} P(y|x)J(y),$$

where we excluded the recurrent state from the sum since after reaching the recurrent state there is no further rewards by definition (2). In vectorial form, $J = r + P'J$ where using (a) we conclude that $J = (I - P')^{-1}r$.

(c) Choose $x \in \{1, \ldots, n\}$. Then,

$$
\begin{aligned}
V(x) =& \mathbb{E}\left[\left(\sum_{k=0}^{\tau-1} r(x_k)\right)^2 \Big| x_0 = x\right] - J(x)^2 \\
=& r(x)^2 + 2r(x)\sum_{y \neq x^*} P(y|x)\mathbb{E}\left[\sum_{k=1}^{\tau-1} r(x_k)\Big| x_1 = y\right] + \\
& \sum_{y \neq x^*} P(y|x)\mathbb{E}\left[\left(\sum_{k=1}^{\tau-1} r(x_k)\right)^2 \Big| x_1 = y\right] - J(x)^2, \\
=& r(x)^2 + 2r(x)\sum_{y \neq x^*} P(y|x)J(y) + \sum_{y \neq x^*} P(y|x)V(y) \\
& + \sum_{y \neq x^*} P(y|x)J(y)^2 - J(x)^2,
\end{aligned}
$$

where in the second equality we took the first term out of the summation, and in the third equality we used the definition of $J$ and $V$. Next, we show that $r(x)^2 + 2r(x)\sum_{y \neq x^*} P(y|x)J(y) + \sum_{y \neq x^*} P(y|x)J(y)^2 - J(x)^2$ is equal to $\rho(x)$:

$$
\begin{aligned}
& r(x)^2 + 2r(x)\sum_{y \neq x^*} P(y|x)J(y) - J(x)^2 + \sum_{y \neq x^*} P(y|x)J(y)^2 \\
& = (r(x) + J(x))(r(x) - J(x)) \\
& \quad + 2r(x)\sum_{y \neq x^*} P(y|x)J(y) + \sum_{y \neq x^*} P(y|x)J(y)^2 \\
& = (r(x) + J(x))\left(-\sum_{y \neq x^*} P(y|x)J(y)\right)
\end{aligned}
$$

$$+ 2r(x) \sum_{y \neq x^*} P(y|x)J(y) + \sum_{y \neq x^*} P(y|x)J(y)^2$$

$$= (r(x) - J(x)) \sum_{y \neq x^*} P(y|x)J(y) + \sum_{y \neq x^*} P(y|x)J(y)^2$$

$$= \sum_{y \neq x^*} P(y|x)J(y)^2 - \left( \sum_{y \neq x^*} P(y|x)J(y) \right)^2 .$$

$\square$

Proposition 3.1 can be used to derive expressions for the gradients w.r.t. $\theta$ of $J$ and $V$. Let $A \circ B$ denote the element-wise product between vectors $A$ and $B$. The gradient expressions are presented in the following lemma.

**Lemma 3.2.** *We have*

$$\nabla J = (I - P')^{-1} \nabla P' J, \qquad (6)$$

*and*

$$\nabla V = (I - P')^{-1} (\nabla \rho + \nabla P' V), \qquad (7)$$

*where*

$$\nabla \rho = \nabla P' J^2 + 2P' (J \circ \nabla J) - 2P' J \circ (\nabla P' J + P' \nabla J). \qquad (8)$$

The proof is a straightforward differentiation of the expressions in Lemma 3.1, and is described in Section A of the supplementary material [3].

We remark that similar equations for the infinite horizon discounted return case were presented by Sobel (1982), in which $I - P'$ is replaced with $I - \beta P$, where $\beta < 1$ is the discount factor. The analysis in (Sobel, 1982) makes use of the fact that $I - \beta P$ is invertible, therefore an extension of their results to the undiscounted case is not immediate.

# 4. Gradient Based Algorithms

In this section we derive gradient based algorithms for solving problems (3) and (4). We present both exact algorithms, which may be practical for small problems, and simulation based algorithms for larger problems. Our algorithms deal with the constraint based on the penalty method, which is described in the following subsection.

## 4.1. Penalty methods

One approach for solving constrained optimization problems (COPs) such as (3) is to transform the COP to an equivalent unconstrained problem, which can be solved using standard unconstrained optimization techniques. These methods, generally known as

*penalty methods*, add to the objective a penalty term for infeasibility, thereby making infeasible solutions suboptimal. Formally, given a COP

$$\max f(x), \quad \text{s.t.} \quad c(x) \leq 0, \qquad (9)$$

we define an unconstrained problem

$$\max f(x) - \lambda g(c(x)), \qquad (10)$$

where $g(x)$ is the *penalty function*, typically taken as $g(x) = (\max(0, x))^2$, and $\lambda > 0$ is the penalty coefficient. As $\lambda$ increases, the solution of (10) converges to the solution of (9), suggesting an iterative procedure for solving (9): solve (10) for some $\lambda$, then increase $\lambda$ and solve (10) using the previous solution as an initial starting point.

In this work we use the penalty method to solve the COP in (3). An alternative approach, which is deferred to future work, is to use *barrier methods*, in which a different penalty term is added to the objective that forces the iterates to remain within the feasible set (Boyd & Vandenberghe, 2004).

## 4.2. Exact Gradient Algorithm

When the MDP transitions are known, the expressions for the gradients in Lemma 3.2 can be immediately plugged into a gradient ascent algorithm for the following penalized objective function of problem (3)

$$f_\lambda = J(x^*) - \lambda g(V(x^*) - b).$$

Let $\alpha_k$ denote a sequence of positive step sizes. Then, a gradient ascent algorithm for maximizing $f_\lambda$ is

$$\theta_{k+1} = \theta_k + \alpha_k (\nabla J(x^*) - \lambda g'(V(x^*) - b) \nabla V(x^*)). \qquad (11)$$

Let us make the following assumption on the smoothness of the objective function and on the set of its local optima. [4]

**Assumption 4.1.** For all $\theta \in \mathbb{R}^{K_\theta}$ and $\lambda > 0$, the objective function $f_\lambda$ has bounded second derivatives. Furthermore, the set of local optima of $f_\lambda$ is countable.

Then, under Assumption 4.1, and suitable conditions on the step sizes, the gradient ascent algorithm (11) can be shown to converge to a locally optimal point of $f_\lambda$.

For the SR optimization problem (4), using the quotient derivative rule for calculating the gradient of $S$,

---

[3] http://tx.technion.ac.il/~avivt/icml12supp.pdf

[4] Note that the smoothness of $J(x^*)$ and $V(x^*)$ may be satisfied by choosing a suitable policy function such as the softmax function.

we obtain the following algorithm

$$\theta_{k+1} = \theta_k + \frac{\alpha_k}{\sqrt{V(x^*)}} \left( \nabla J(x^*) - \frac{J(x^*)}{2V(x^*)} \nabla V(x^*) \right),$$ (12)

which can be shown to converge under similar conditions to a locally optimal point of (4).

When the state space is large, or when the model is not known, computation of the gradients using equations (6) and (7) is not feasible. In these cases, we can use simulation to obtain unbiased estimates of the gradients, as we describe in the next section, and perform a *stochastic* gradient ascent.

### 4.3. Simulation based optimization

When a simulator of the MDP dynamics is available, it is possible to obtain unbiased estimates of the gradients $\nabla J$ and $\nabla V$ from a sample trajectory between visits to the recurrent state. The technique is called the *likelihood ratio* method, and it underlies all policy gradient algorithms (Baxter & Bartlett, 2001; Marbach & Tsitsiklis, 1998). The following lemma gives the necessary gradient estimates for our case.

**Lemma 4.2.** *We have*

$$\nabla J(x) = \mathbb{E}[R_0^{\tau-1} \nabla \log P \left( x_0^{\tau-1} \right) | x_0 = x],$$

*and*

$$\nabla V(x) = \mathbb{E}[\left( R_0^{\tau-1} \right)^2 \nabla \log P \left( x_0^{\tau-1} \right) | x_0 = x] - 2J(x) \nabla J(x),$$

*where the expectation is over trajectories.*

The proof is given in Section B of the supplementary material.

Given an observed trajectory $x_0^{\tau-1}, u_0^{\tau-1}, r_0^{\tau-1}$, and using Lemma 4.2 we devise the estimator $\hat{\nabla} J(x^*) \triangleq R_0^{\tau-1} \nabla \log P \left( x_0^{\tau-1} \right)$ which is an unbiased estimator of $\nabla J(x^*)$. Furthermore, using the Markov property of the state transition and the fact that the only dependance on $\theta$ is in the policy $\mu_\theta$, the term $\nabla \log P \left( x_0^{\tau-1} \right)$ can be reduced to

$$\nabla \log P \left( x_0^{\tau-1} \right) = \sum_{k=0}^{\tau-1} \nabla \log \mu_\theta \left( u_k | x_k \right),$$

making the computation of $\hat{\nabla} J(x^*)$ from an observed trajectory straightforward. Assume for the moment that we know $J(x^*)$ and $V(x^*)$. Then $\hat{\nabla} V(x^*) \triangleq (R_0^{\tau-1})^2 \nabla \log P \left( x_0^{\tau-1} \right) - 2J(x^*) \hat{\nabla} J(x^*)$ is an unbiased estimate of $\nabla V(x^*)$, and plugging $\hat{\nabla} V$ and $\hat{\nabla} J$ in (11) gives a proper stochastic gradient ascent algorithm.

Unfortunately, we cannot calculate $J(x^*)$ exactly without knowing the model, and obtaining an unbiased estimate of $J(x)\nabla J(x)$ from a single trajectory is impossible (for a similar reason that the variance of a random variable cannot be estimated from a single sample of it). We overcome this difficulty by using a two time-scale algorithm, where estimates of $J$ and $V$ are calculated on the fast time scale, and $\theta$ is updated on a slower time scale.

The algorithm updates the parameters every episode, upon visits to the recurrent state $x^*$. Let $\tau^k$ where $k = 0, 1, 2, \dots$ denote the times of these visits. To ease notation, we also define $x^k = (x_{\tau_{k-1}}, \dots, x_{\tau_k - 1})$ and $R^k = \sum_{t=\tau_{k-1}}^{\tau_k - 1} r_t$ to be the trajectories and accumulated rewards observed between visits, and denote $z^k \triangleq \nabla \log P(x^k)$ to be the likelihood ratio derivative. The simulation based algorithm for the constrained optimization problem (3) is

$$\tilde{J}_{k+1} = \tilde{J}_k + \alpha_k \left( R^k - \tilde{J}_k \right)$$

$$\tilde{V}_{k+1} = \tilde{V}_k + \alpha_k \left( (R^k)^2 - \tilde{J}_k^2 - \tilde{V}_k \right)$$

$$\theta_{k+1} = \theta_k + \beta_k \left( R^k - \lambda g' \left( \tilde{V}_k - b \right) \left( (R^k)^2 - 2\tilde{J}_k \right) \right) z^k,$$ (13)

where $\alpha_k$ and $\beta_k$ are positive step sizes. Similarly, for optimizing the SR (4), we change the update rule for $\theta$ to

$$\theta_{k+1} = \theta_k + \frac{\beta_k}{\sqrt{\tilde{V}_k}} \left( R^k - \frac{\tilde{J}_k (R^k)^2 - 2R^k \tilde{J}_k^2}{2\tilde{V}_k} \right) z^k.$$ (14)

In the next theorem we prove that algorithm (13) converges almost surely to a locally optimal point of the corresponding objective function. The proof for Algorithm (14) is essentially the same and thus omitted. For notational clarity, throughout the remainder of this section, the dependence of $J(x^*)$ and $V(x^*)$ on $\theta$ is made explicit using a subscript.

**Theorem 4.3.** *Consider algorithm* (13), *and let Assumptions 2.1, 2.2, and 4.1 hold. If the step size sequences satisfy* $\sum_k \alpha_k = \sum_k \beta_k = \infty$, $\sum_k \alpha_k^2, \sum_k \beta_k^2 < \infty$, *and* $\lim_{k \to \infty} \frac{\beta_k}{\alpha_k} = 0$, *then almost surely*

$$\lim_{k \to \infty} \nabla \left( J_{\theta_k}(x^*) - \lambda g \left( V_{\theta_k}(x^*) - b \right) \right) = 0.$$ (15)

*Proof.* (sketch) The proof relies on representing Equation (13) as a stochastic approximation with two time-scales (Borkar, 1997), where $\tilde{J}_k$ and $\tilde{V}_k$ are updated on a fast schedule while $\theta_k$ is updated on a slow schedule. Thus, $\theta_k$ may be seen as quasi-static w.r.t. $\tilde{J}_k$ and $\tilde{V}_k$, suggesting that $\tilde{J}_k$ and $\tilde{V}_k$ may be associated with the following ordinary differential equations (ODE)

$$\dot{J} = \mathbb{E}_\theta[B|x_0 = x^*] - J,$$
$$\dot{V} = \mathbb{E}_\theta[B^2|x_0 = x^*] - J^2 - V. \qquad (16)$$

For each $\theta$, the ODE (16) can be solved analytically to yield $J(t) = J^\infty + c_1 e^{-t}$ and $V(t) = V^\infty - 2J^\infty c_1 t e^{-t} + c_1^2 e^{-2t} + c_2 e^{-t}$, where $c_1$ and $c_2$ are constants, and $\{J^\infty, V^\infty\}$ is a globally asymptotically stable fixed point which satisfies

$$J^\infty = J_\theta(x^*), \quad V^\infty = V_\theta(x^*). \qquad (17)$$

In turn, due to the timescale difference, $\tilde{J}_k$ and $\tilde{V}_k$ in the iteration for $\theta_k$ may be replaced with their stationary limit points $J^\infty$ and $V^\infty$, suggesting the following ODE for $\theta$

$$\dot{\theta} = \nabla \left( J_\theta(x^*) - \lambda g \left( V_\theta(x^*) - b \right) \right). \qquad (18)$$

Under Assumption 4.1, the set of stable fixed point of (18) is just the set of locally optimal points of the objective function $f_\lambda$. Let $\mathcal{Z}$ denote this set, which by Assumption 4.1 is countable. Then, by Theorem 5 in Leslie & Collins, 2002 (which is extension of Theorem 1.1 in Borkar, 1997), $\theta_k$ converges to a point in $\mathcal{Z}$ almost surely. □

## 5. Experiments

In this section we apply the simulation based algorithms of Section 4 to a portfolio management problem, where the available investment options include both liquid and non-liquid assets. In the interest of understanding the performance of the different algorithms, we consider a rather simplistic model of the corresponding financial problem. We emphasize that dealing with richer models requires no change in the algorithms.

We consider a portfolio that is composed of two types of assets. A liquid asset (e.g., short term T-bills), which has a fixed interest rate $r_l$ but may be sold at every time step $t = 1, \ldots, T$, and a non-liquid asset (e.g., low liquidity bonds or options) that has a time dependent interest rate $r_{nl}(t)$, yet may be sold only after a maturity period of $N$ steps. In addition, the non-liquid asset has some risk of not being paid (i.e., a default) with a probability $p_{risk}$. A common investment strategy in this setup is *laddering*–splitting the investment in the non-liquid assets to chunks that are reinvested in regular intervals, such that a regular cash flow is maintained. In our model, at each time step the investor may change his portfolio by investing a fixed fraction $\alpha$ of his total available cash in a non-liquid asset. Of course, he can only do that when he has at least $\alpha$ invested in liquid assets, otherwise he has to wait until enough non-liquid assets mature. In addition, we assume that at each $t$ the interest rate $r_{nl}(t)$ takes one of two values - $r_{nl}^{high}$ or $r_{nl}^{low}$, and the transitions between these values occur stochastically with switching probability $p_{switch}$. The state of the model at each time step is represented by a vector $x(t) \in \mathbb{R}^{N+2}$, where $x_1 \in [0, 1]$ is the fraction of the investment in liquid assets, $x_2, \ldots, x_{N+1} \in [0, 1]$ is the fraction in non-liquid assets with time to maturity of $1, \ldots, N$ time steps, respectively, and $x_{N+2}(t) = r_{nl}(t) - \mathbb{E}[r_{nl}(t)]$. At time $t = 0$ we assume that all investments are in liquid assets, and we denote $x^* = x(t = 0)$. The binary action at each step is determined by a stochastic policy, with probability $\mu_\theta(x) = \epsilon + (1 - 2\epsilon) / \left(1 + e^{-\theta x}\right)$ of investing in a non-liquid asset. Note that this 'ε-constrained' softmax policy comes to satisfy Assumption 2.3. Our reward is just the logarithm of the return from the investment (which is additive at each step). The dynamics of the investment chunks are illustrated in Figure 2.

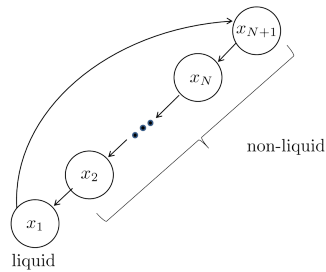

*Figure 2.* Dynamics of the investment.

We optimized the policy parameters using the simulation based algorithms of Section 4 with three different performance criteria: (a) Average reward: $\max J(x^*)$, (b) Variance constrained reward $\max J(x^*)$ s.t. $V(x^*) \le b$, and (c) the SR $\max J(x^*) \sqrt{V(x^*)}$. Figure 3 shows the distribution of the accumulated reward. As anticipated, the policy for criterion (a) was risky, and yielded higher gain than the policy for the variance constrained criterion (b). Interestingly, maximizing the SR resulted in a very conservative policy, that almost never invested in the non-liquid asset. The parameters for the experiments are detailed in the supplementary material, Section C.

## 6. Conclusion

This work presented a novel algorithmic approach for RL with variance related risk criteria, a subject that while being important for many applications, has been notoriously known to pose significant algorithmic challenges. Since getting to an optimal solution seems hard even when the model is known, we adopted a gradient based approach that achieves local optimality.

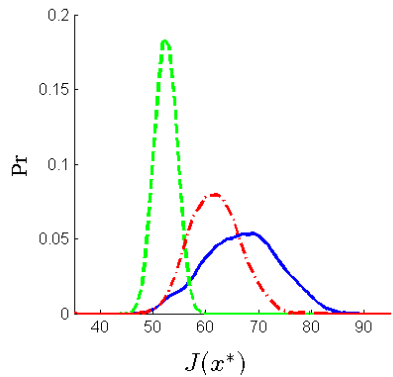

*Figure 3.* Distribution of the accumulated reward. Solid line: corresponds to the policy obtained by maximizing total reward. Dash-dotted line: maximizing total reward s.t. variance less than 20. Dashed line : maximize the SR.

A few issues are in need of further investigation. First, we note a possible extension to other risk measures such as the percentile criterion (Delage & Mannor, 2010). This will require a result reminiscent to Proposition 3.1 that would allow us to drive the optimization. Second, we could consider variance in the optimization process to improve convergence time in the style of *control variates*. Policy gradient algorithms are known to suffer from high variance when the recurrent state in not visited frequently. One technique for dealing with this difficulty is by using control variates (Greensmith et al., 2004). Imposing a variance constraint as described in this work also acts along this direction, and may in fact improve performance of such algorithms even if variance is not part of the criterion we are optimizing. Third, policy gradients are just one family of algorithms we can consider. It would be interesting to see if a temporal-difference style algorithm can be developed for the risk measures considered here. Lastly, we note that experimentally, maximizing the SR resulted in a very risk averse behavior. This interesting phenomenon deserves more research. It suggests that it might be more prudent to consider other risk measures instead of the SR.

## Acknowledgements

The research leading to these results has received funding from the European Unions Seventh Framework Programme (FP7/2007-2013) under PASCAL2 (PUMP PRIMING) grant agreement no. 216886 and under a Marie Curie Reintegration Fellowship (IRG) grant agreement no. 249254.